\documentclass[10pt,twocolumn,letterpaper]{article}
\usepackage{authblk}
\usepackage[accsupp]{axessibility}
\usepackage{cvpr}              

\usepackage[dvipsnames]{xcolor}

\usepackage[symbol]{footmisc}

\definecolor{cvprblue}{rgb}{0.21,0.49,0.74}
\usepackage[pagebackref,breaklinks,colorlinks,citecolor=cvprblue]{hyperref}
\usepackage{graphicx}
\usepackage{tikz}
\usepackage{amsmath}
\usepackage{amssymb}
\usepackage{booktabs}

\usepackage{times}
\usepackage{epsfig}
\usepackage{graphicx}
\usepackage{amsmath}
\usepackage{amssymb}

\usepackage{booktabs}
\usepackage{tabularx,ragged2e,booktabs,caption}
\usepackage{multicol,tabularx,capt-of}
\usepackage{hhline}
\usepackage{multirow}
\usepackage{physics}
\usepackage{pbox}
\def\paperID{40} 
\def\confName{CVPR}
\def\confYear{2024}

\makeatletter
\renewcommand\AB@affilsepx{, \protect\Affilfont}
\makeatother

\title{Salient Object-Aware Background Generation using Text-Guided Diffusion Models}

\author[1]{Amir Erfan Eshratifar}
\author[1]{Jo\~ao V. B. Soares}
\author[1]{Kapil Thadani}
\author[2,*]{\authorcr Shaunak Mishra}
\author[2*]{Mikhail Kuznetsov}
\author[3,*]{Yueh-Ning Ku}
\author[1]{Paloma de Juan}
\affil[1]{Yahoo Research, USA}
\affil[2]{Amazon, USA}
\affil[3]{ByteDance, USA}

\begin{document}
\maketitle
\newcommand{\rulesep}{\unskip\ \vrule\ }

\newcommand{\columnname}[1]
{\makebox[.29\columnwidth][c]{#1}}

\newcommand{\prompttext}[1]
{\makebox[.59\columnwidth][c]{\footnotesize #1}}

\thispagestyle{empty}

\begin{abstract}
Generating background scenes for salient objects plays a crucial role across various domains including creative design and e-commerce, as it enhances the presentation and context of subjects by integrating them into tailored environments. Background generation can be framed as a task of text-conditioned outpainting, where the goal is to extend image content beyond a salient object's boundaries on a blank background.
Although popular diffusion models for text-guided inpainting can also be used for outpainting by mask inversion, they are trained to fill in missing parts of an image rather than to place an object into a scene. Consequently, when used for background creation, inpainting models frequently extend the salient object's boundaries and thereby change the object’s identity, which is a phenomenon we call ``object expansion." This paper introduces a model for adapting inpainting diffusion models to the salient object outpainting task using Stable Diffusion and ControlNet architectures. We present a series of qualitative and quantitative results across models and datasets, including a newly proposed metric to measure object expansion that does not require any human labeling. Compared to Stable Diffusion 2.0 Inpainting, our proposed approach reduces object expansion by 3.6$\times$ on average with no degradation in standard visual metrics across multiple datasets.
\end{abstract}

\footnote[1]{Work done at Yahoo Research.}

\begin{figure*}
\begin{center}

    \columnname{Salient Object}\hfill
    \columnname{Stable Inpainting}\hfill
    \columnname{Ours}\hfill
    \columnname{Stable Inpainting}\hfill
    \columnname{Ours}\hfill
    \columnname{Stable Inpainting}\hfill
    \columnname{Ours}\\

    \includegraphics[width=.29\columnwidth]{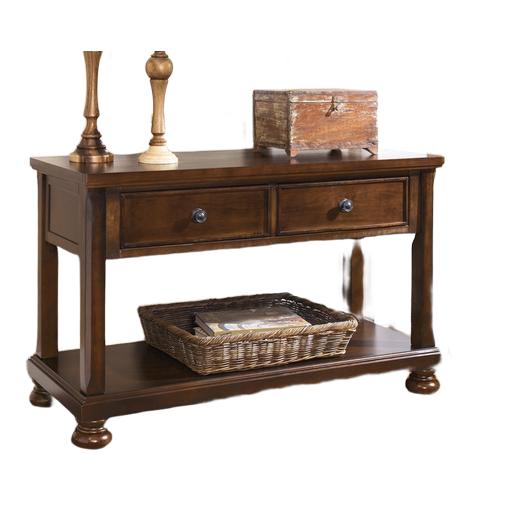}\hfill \rulesep
    \includegraphics[width=.29\columnwidth]{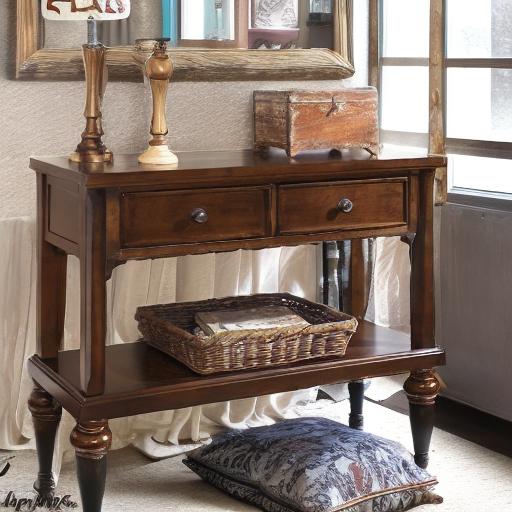}\hfill
    \includegraphics[width=.29\columnwidth]{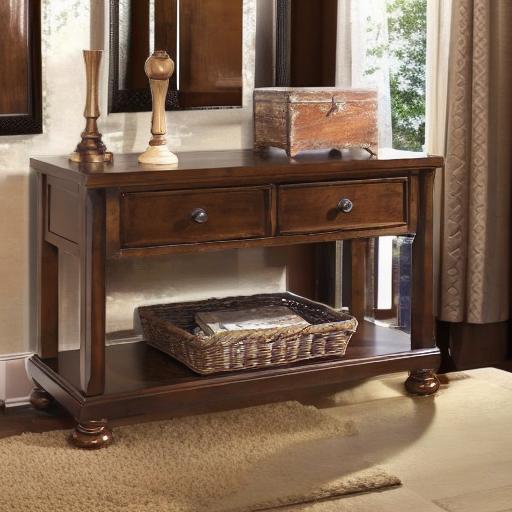}\hfill \rulesep
    \includegraphics[width=.29\columnwidth]{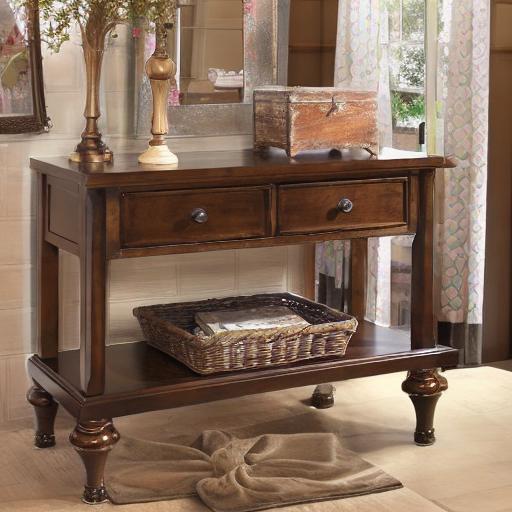}\hfill
    \includegraphics[width=.29\columnwidth]{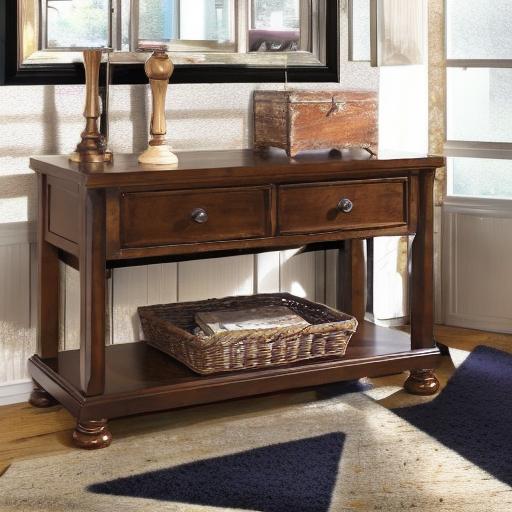}\hfill \rulesep
    \includegraphics[width=.29\columnwidth]{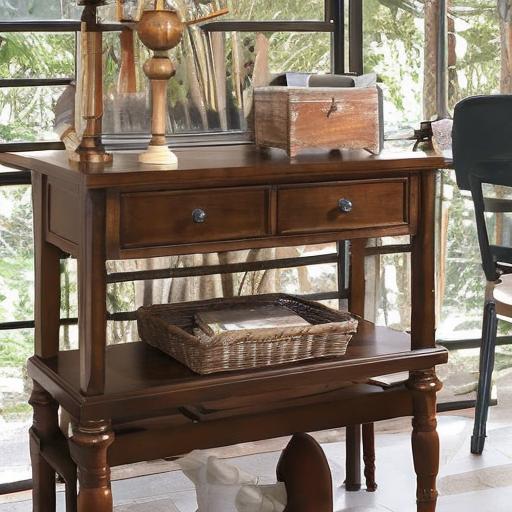}\hfill
    \includegraphics[width=.29\columnwidth]{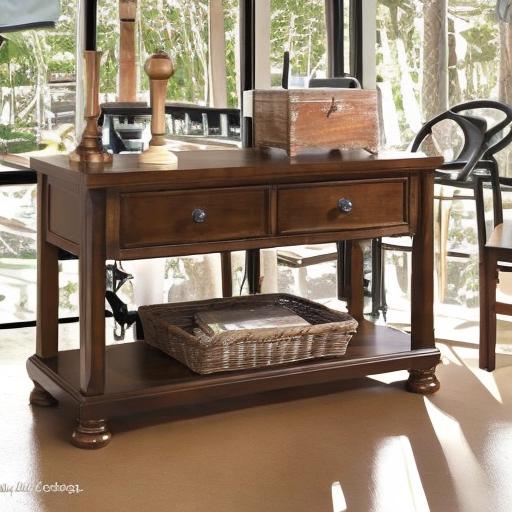}\hfill
    \columnname{}\hfill
    \prompttext{``a table in a room''}\hfill
    \prompttext{``a table in a room''}\hfill
    \prompttext{``a table in a living room''}\hfill
    \\[\smallskipamount]
    
    \includegraphics[width=.29\columnwidth]{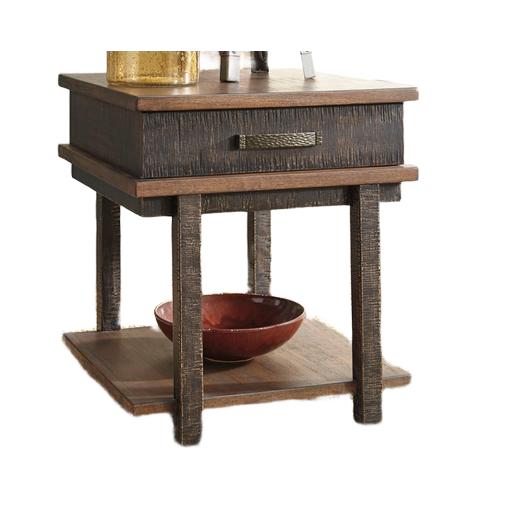}\hfill \rulesep
    \includegraphics[width=.29\columnwidth]{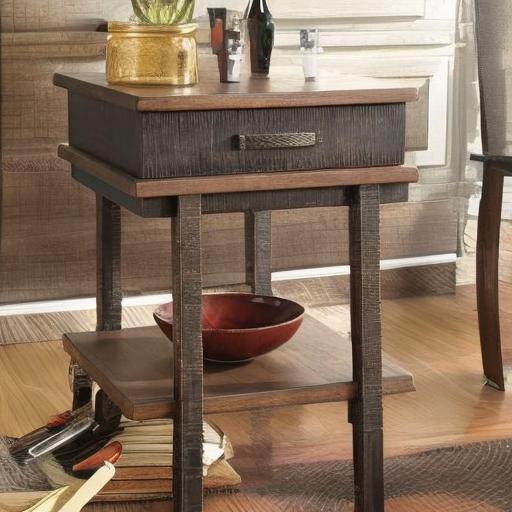}\hfill
    \includegraphics[width=.29\columnwidth]{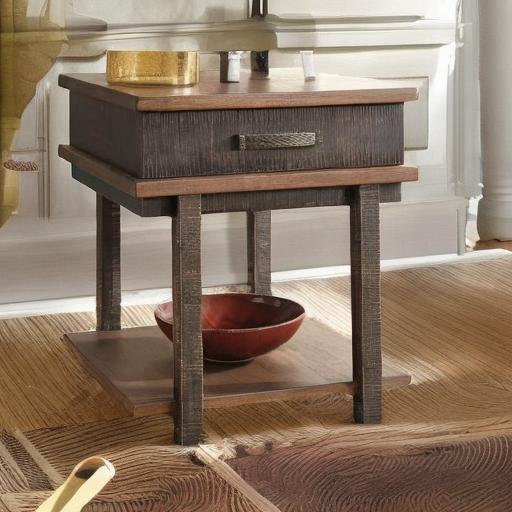}\hfill \rulesep
    \includegraphics[width=.29\columnwidth]{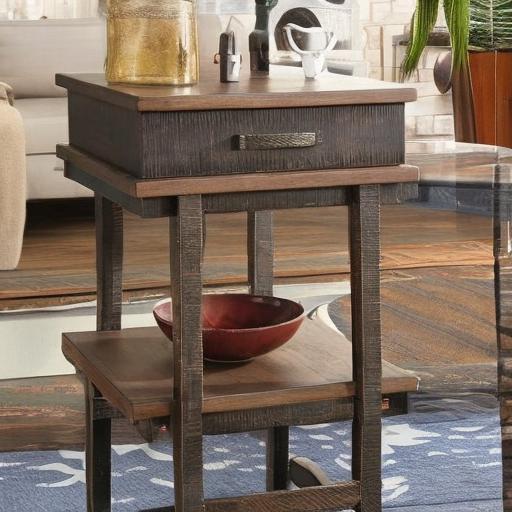}\hfill
    \includegraphics[width=.29\columnwidth]{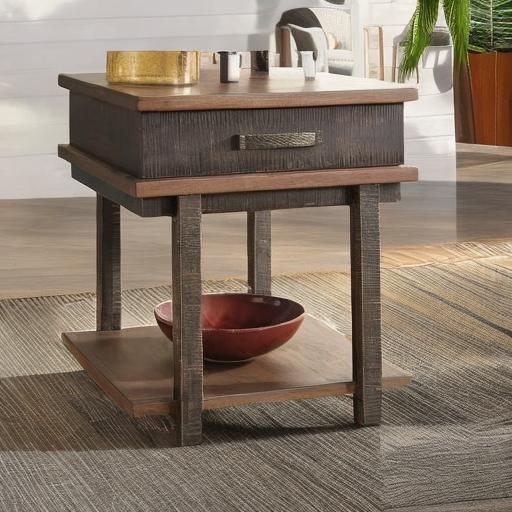}\hfill \rulesep
    \includegraphics[width=.29\columnwidth]{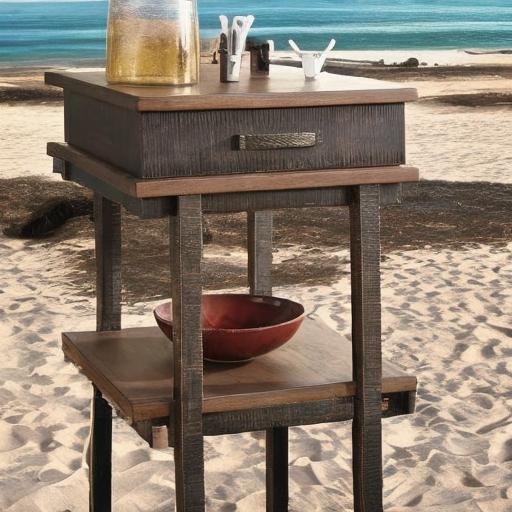}\hfill
    \includegraphics[width=.29\columnwidth]{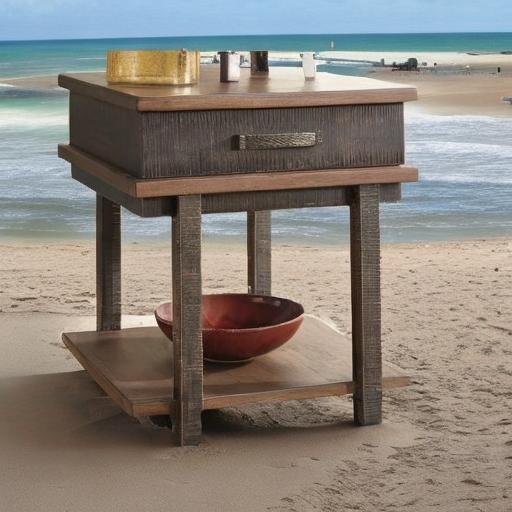}\hfill
    \columnname{}\hfill
    \prompttext{``a table in a room''}\hfill
    \prompttext{``a table in a room''}\hfill
    \prompttext{``a table on a beach''}\hfill
    \\[\smallskipamount]
    \includegraphics[width=.29\columnwidth]{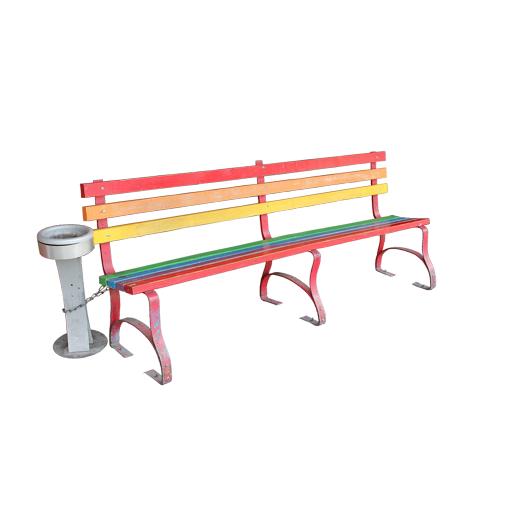}\hfill \rulesep
    \includegraphics[width=.29\columnwidth]{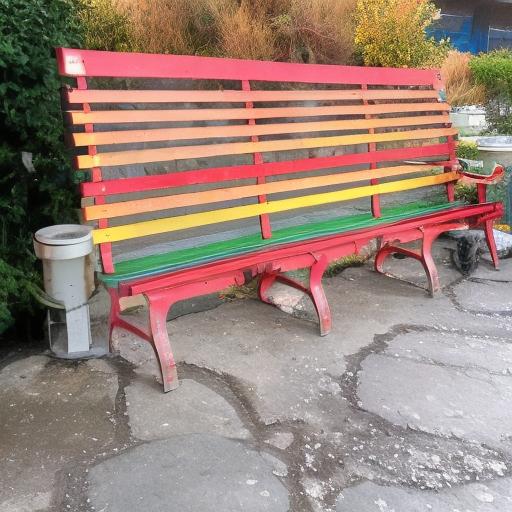}\hfill
    \includegraphics[width=.29\columnwidth]{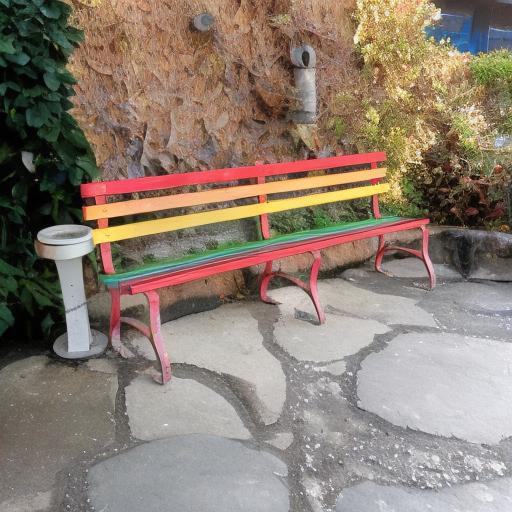}\hfill \rulesep
    \includegraphics[width=.29\columnwidth]{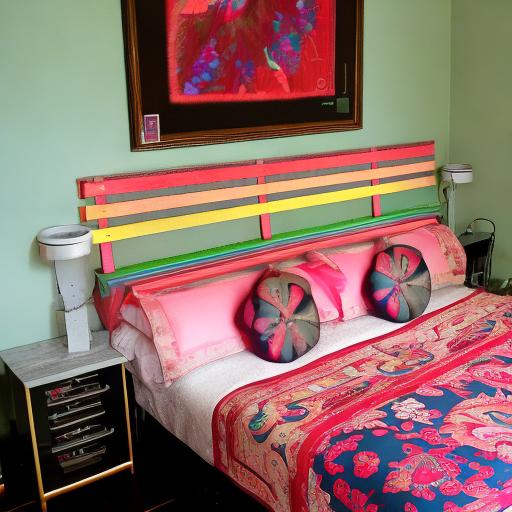}\hfill
    \includegraphics[width=.29\columnwidth]{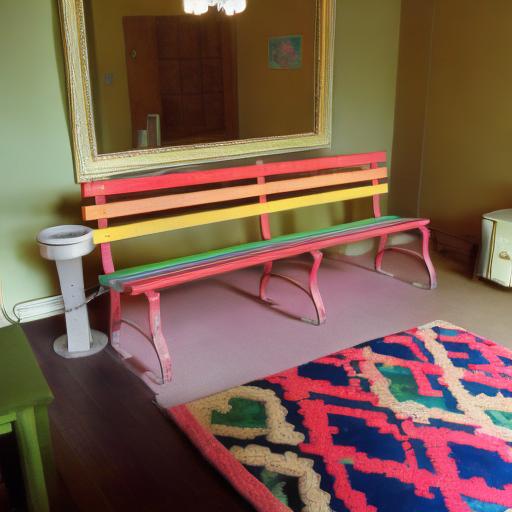}\hfill \rulesep
    \includegraphics[width=.29\columnwidth]{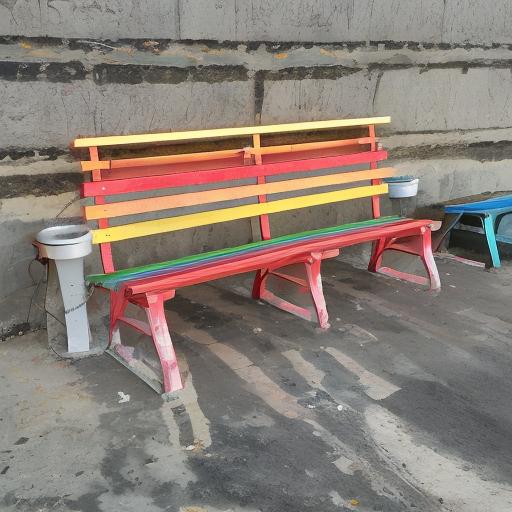}\hfill
    \includegraphics[width=.29\columnwidth]{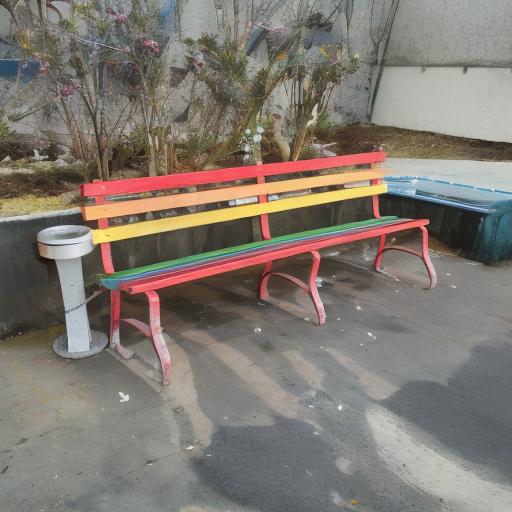}\hfill
    \columnname{}\hfill
    \prompttext{``a bench in a park''}\hfill
    \prompttext{``a bench in a room''}\hfill
    \prompttext{``a bench in a place''}\hfill
    \\[\smallskipamount]
    \includegraphics[width=.29\columnwidth]{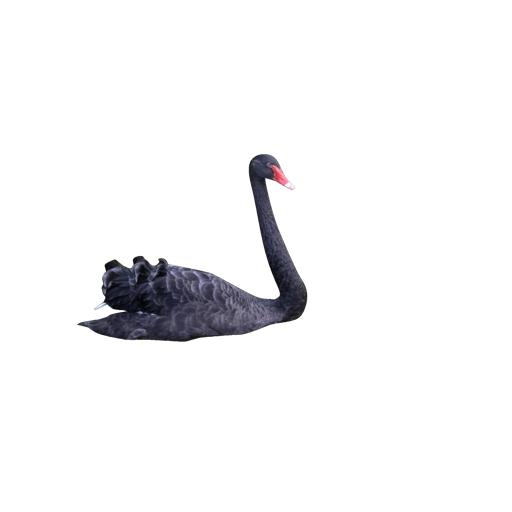}\hfill \rulesep
    \includegraphics[width=.29\columnwidth]{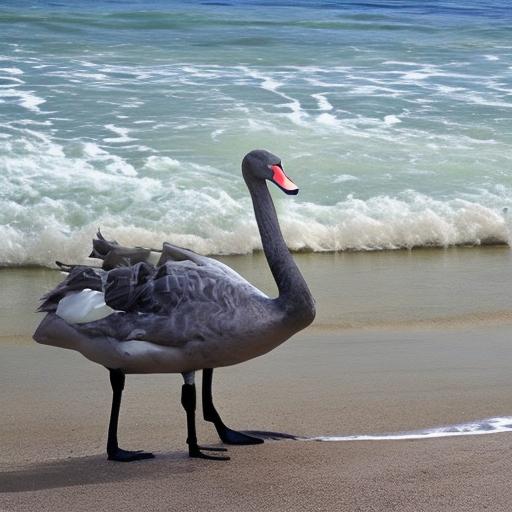}\hfill  
    \includegraphics[width=.29\columnwidth]{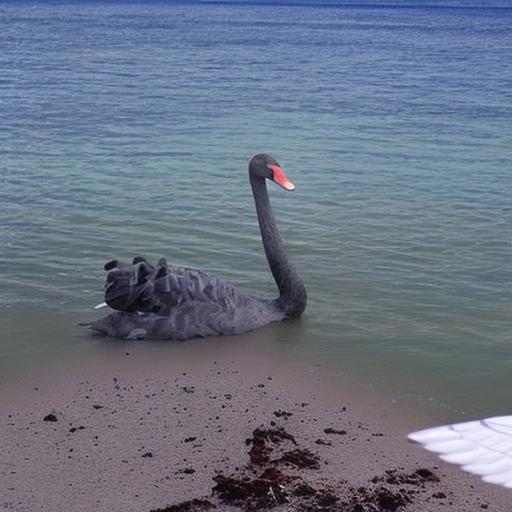}\hfill   \rulesep
    \includegraphics[width=.29\columnwidth]{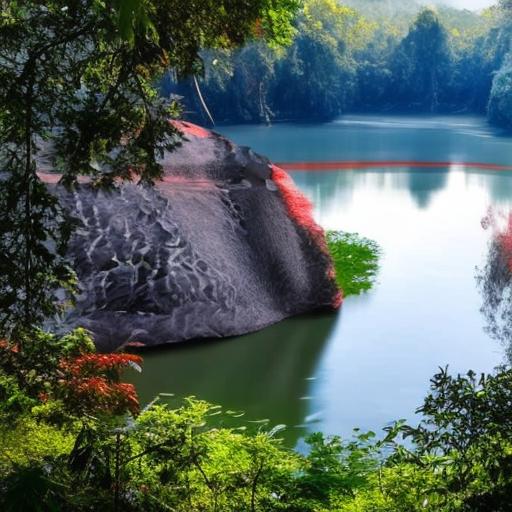}\hfill  
    \includegraphics[width=.29\columnwidth]{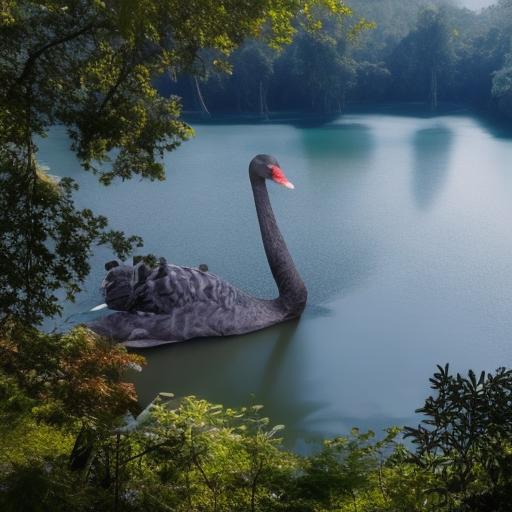}\hfill  \rulesep
    \includegraphics[width=.29\columnwidth]{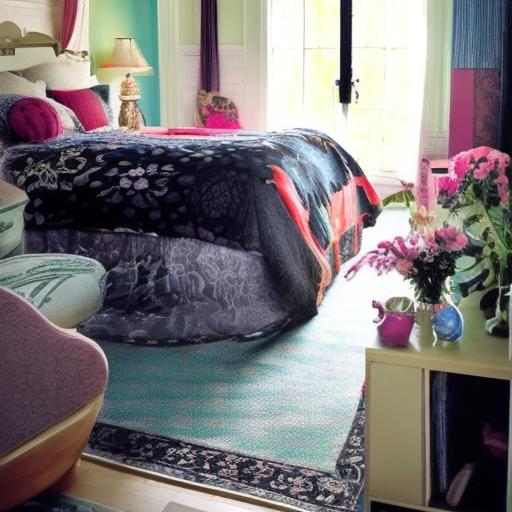}\hfill 
    \includegraphics[width=.29\columnwidth]{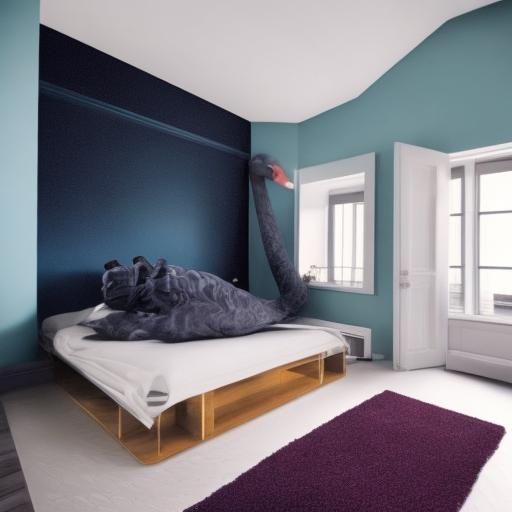}\hfill
     \columnname{}\hfill
    \prompttext{``a swan on a beach''}\hfill
    \prompttext{``a swan in a jungle''}\hfill
    \prompttext{``a swan in a room''}\hfill

\end{center}
   \caption{Examples of outpainting a salient object (leftmost column) using the Stable Inpainting 2.0 (SI2) model (columns 2, 4, 6 from left) and using our proposed model (columns 3, 5, 7 from left). The images in each paired column (2 \& 3, 4 \& 5, 6 \& 7) are generated using the same seed and prompt, but one uses SI2, and the other uses our model. Objects are often expanded using the SI2 model, which may catastrophically change the object's identity. For example, the legs of the tables are expanded in the first two rows; in the third row, a bench is transformed into a bed; in the last row, a swan is blended into a rock and a bed. 
   }
\label{fig:salient_examples}
\end{figure*}

\begin{figure}[h]
    \begin{subfigure}[c]{0.25\columnwidth}
        \caption{Input image}
        \includegraphics[width=0.93\columnwidth]{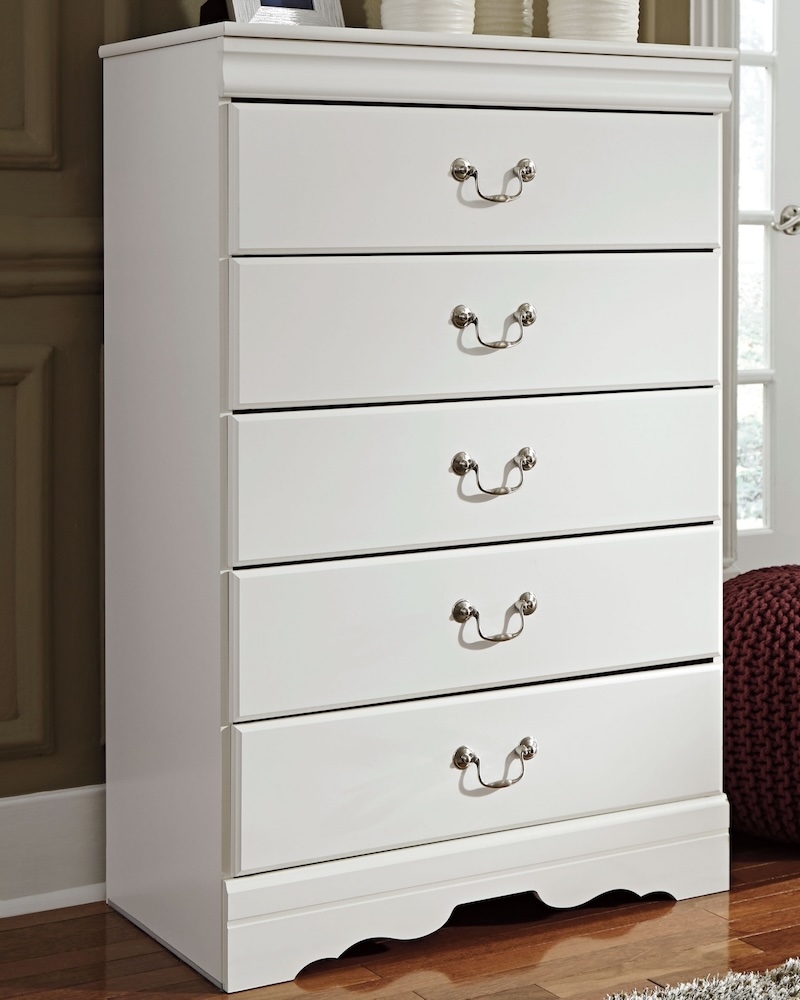}
        \label{img1}
    \end{subfigure}
    \begin{subfigure}[c]{0.75\columnwidth}
        \caption{RunwayML's generated backgrounds}
        \includegraphics[width=0.31\columnwidth]{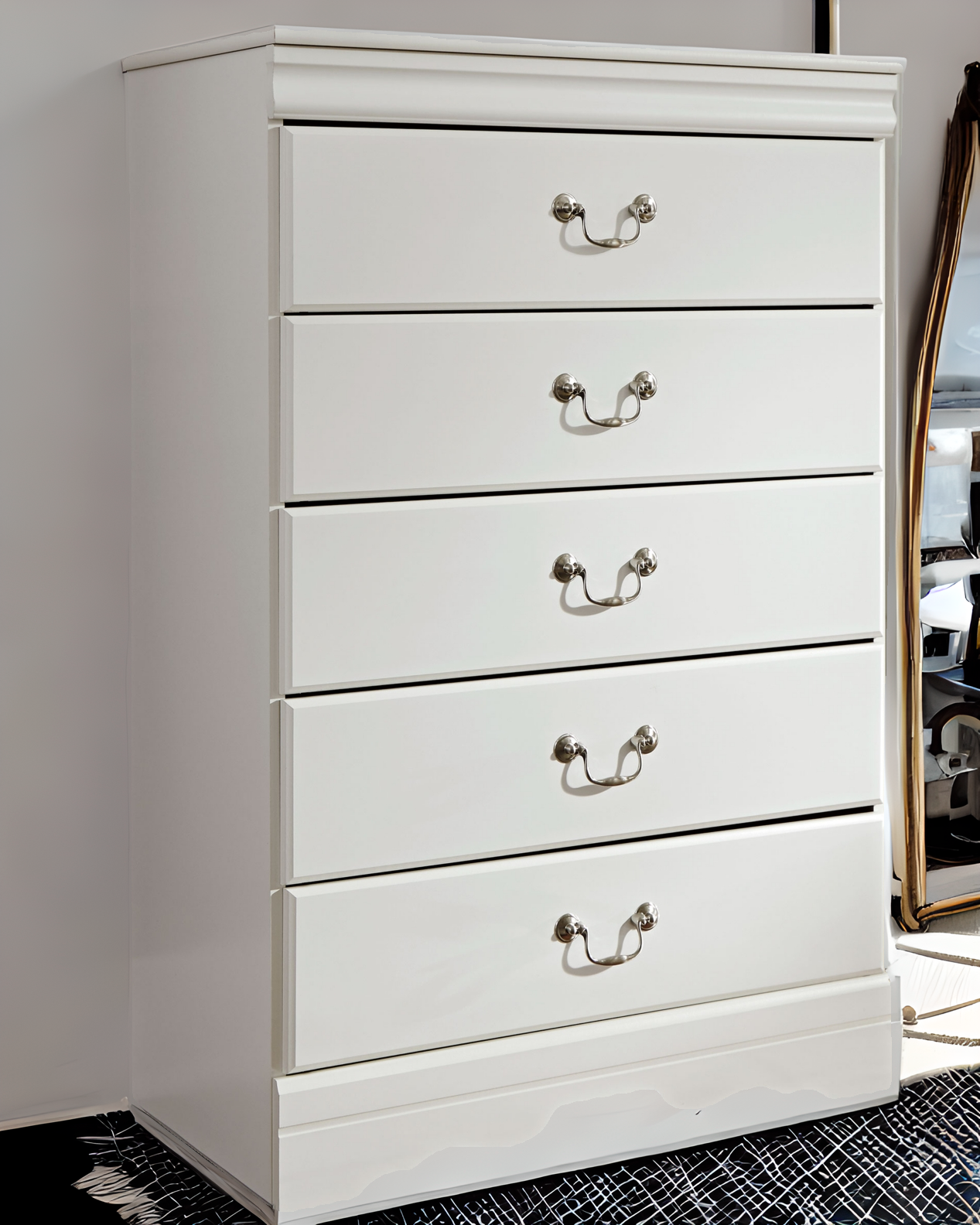}
        \includegraphics[width=0.31\columnwidth]{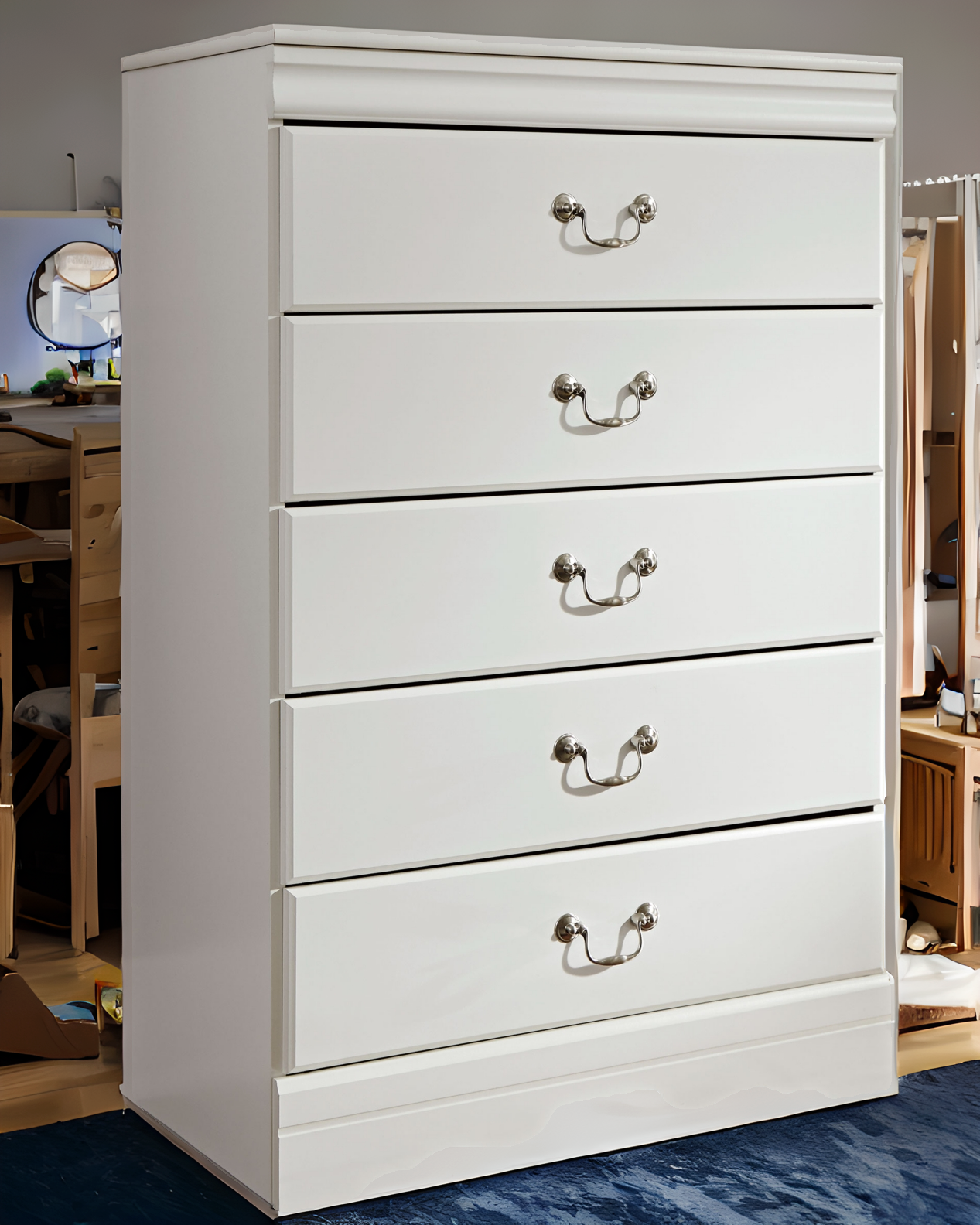}
        \includegraphics[width=0.31\columnwidth]{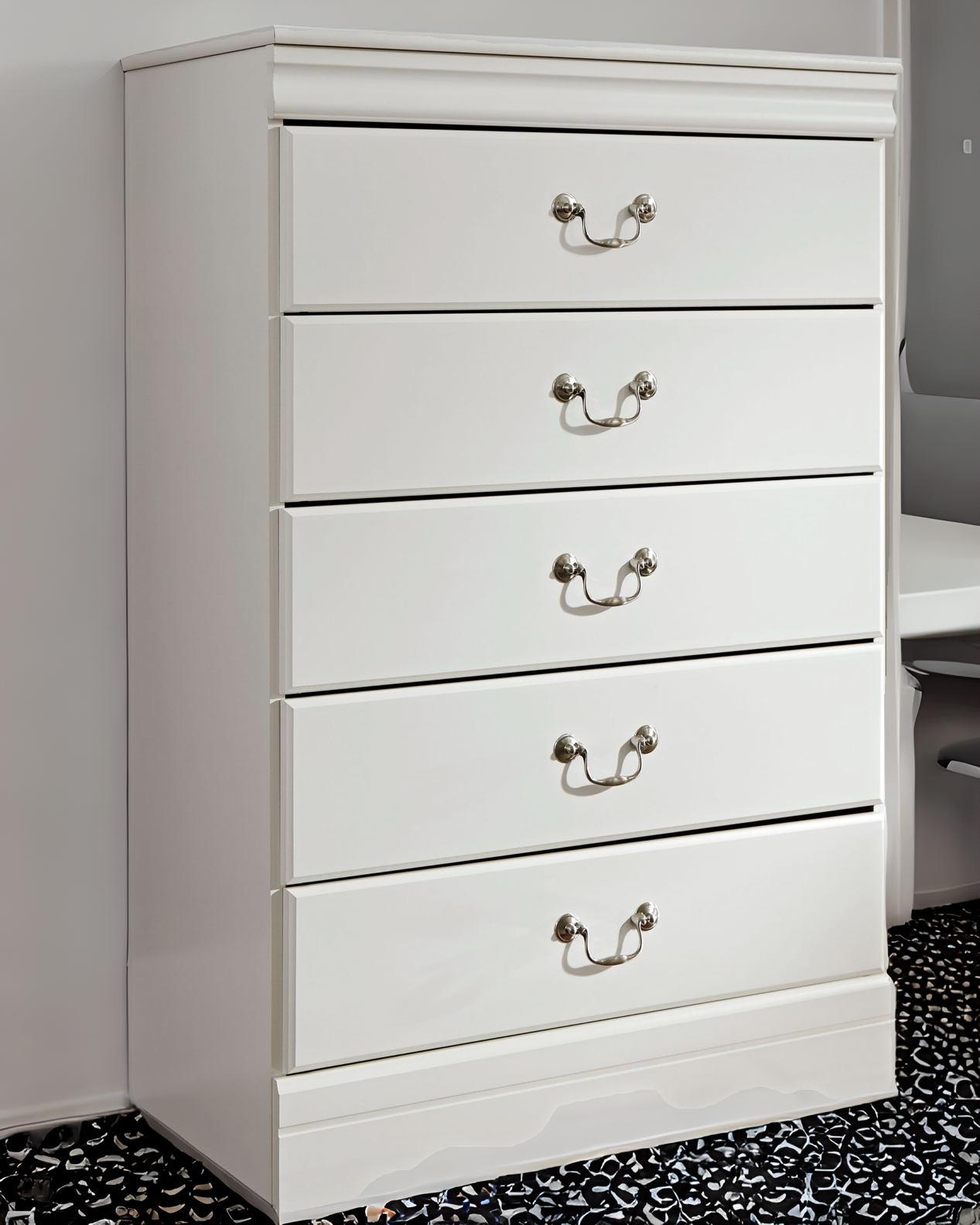}
        \label{img2}
    \end{subfigure}
    \caption{Significant object expansion is seen at the bottom of the white dresser with RunwayML's Background Remix, a popular commercial tool. These examples are generated with the prompt of ``a modern room.''}
    
    \label{fig:runwayml_background_remix}
\end{figure}
\section{Introduction}
Image outpainting, also known as image extrapolation or extension, has been a longstanding challenge within computer vision. Prior image outpainting techniques relied upon retrieval and stitching methods using image patches, or learning-based methods \cite{5473074, 10.1145/2661229.2661278, 6618999, 10.1007/978-3-030-58529-7_41, 9008290, 8953261}.
The new wave of generative image models~\cite{Rombach2021HighResolutionIS,Ramesh2022HierarchicalTI,Saharia2022PhotorealisticTD}
has been adapted to also solve the outpainting task, representing a breakthrough in image quality and adding controllability via text prompts and other control inputs. Our work focuses specifically on \emph{salient object outpainting}, an outpainting problem that involves generating a natural and coherent background for a salient object, while optionally conditioned on a text prompt, as shown in Figure \ref{fig:salient_examples}. 

Given an object, humans can readily imagine its empirical context by 
relating objects to their context in daily life while also being able to imagine them in unconventional settings, such as a swan in a bedroom, as depicted in Figure \ref{fig:salient_examples}. There are many potential applications for the salient object outpainting problem we study here, such as generating backgrounds for products in online advertising, film-making, creative design, and augmented reality. Object outpainting is much more challenging than usual image completion tasks like inpainting and outpainting for two reasons: (i) the object and background contents may not be related to each other, (ii) to generate a background constrained by the salient object, the model needs to understand the correlations within the scene at a semantic level.

Recently, diffusion models \cite{NEURIPS2020_4c5bcfec, pmlr-v37-sohl-dickstein15} such as Latent Diffusion Models \cite{Rombach2021HighResolutionIS}, unCLIP \cite{Ramesh2022HierarchicalTI}, and Imagen \cite{Saharia2022PhotorealisticTD} have shown outstanding results in text-to-image generation. An early approach for adapting them to the inpainting task consisted of replacing the random noise in the fixed portion of the image with a noisy version of itself during the diffusion reverse process \cite{Lugmayr2022RePaintIU}; however, the model's inability to observe the global context during sampling led to unsatisfactory samples \cite{Nichol2021GLIDETP}. GLIDE \cite{Nichol2021GLIDETP} and Stable Inpainting (SI)~\cite{Rombach2021HighResolutionIS} improved upon this by using the masked image as extra conditioning information to the reverse diffusion process. They trained their models using randomly generated masks to specify what portion of the image to inpaint;
however, the masks were randomly placed and had circular, square, or highly irregular shapes which are rarely ever seen in real-world inpainting scenarios.
To better mimic real-world inpainting masks, later works~\cite{10.1007/978-3-030-58529-7_1, 10.1007/978-3-031-19787-1_16} propose masks that follow object shapes obtained from various segmentation datasets. To ensure that salient objects are not masked out, they subtract the portion of some masks that correspond to salient objects in the training image.
However, these masks can be from any object, resulting in small masks relative to the image size. Thus, despite these improvements, inpainting models are primarily trained to fill in missing parts of an image rather than synthesize complete backgrounds conditioned on salient objects.

In practice, we observe that when 
such inpainting models are used for background generation, they often ignore the salient object’s original boundaries and modify or recharacterize the object, as shown in Figure \ref{fig:salient_examples} for the Stable Inpainting 2.0 model (SI2)~\cite{stableinpaintingtwo,Rombach2021HighResolutionIS}. We call this phenomenon \emph{object expansion}. As shown in Figure \ref{fig:runwayml_background_remix}, even popular commercial tools for background generation are prone to this limitation.
To quantify object expansion, we propose an automated metric that avoids the need for any human labeling. We also propose a solution to object expansion using ControlNet~\cite{zhang2023adding} to maintain object boundaries. ControlNet was introduced to control large text-to-image models with extra input conditions like edge maps, segmentation maps, key points, etc. We utilize the mask of the salient object as a new input condition to address expansion. Although ControlNet was initially designed for standard text-to-image diffusion models, we modify its architecture to be compatible with diffusion-based inpainting. 

Though our approach could also be applied to background generation for non-salient objects, we train our model specifically for salient object outpainting for two reasons.
First, we can use readily available manually annotated salient object detection datasets to train our salient object outpainting model. If we were to tackle background generation for non-salient objects, we would have to train our model with panoptic or instance segmentation datasets, whose masks, unfortunately, are not pixel-perfect, containing noisy labels that may not help reduce the object expansion problem. Second, a notable application for background generation is e-commerce, where personalized and aesthetically pleasant backgrounds can be generated for products that would be salient objects in the final images.

%
Our main contributions are:
\begin{itemize}
  \item A novel 
  study of diffusion-based inpainting models applied to salient object-aware background generation.
  \item A characterization of the object expansion problem when inpainting models are applied for background generation, as well as a measure for quantifying it.
  \item An architecture based on ControlNet for adapting diffusion-based inpainting models to salient object-aware background generation.\footnote{The codes and model checkpoints will be available at \url{https://github.com/yahoo/photo-background-generation}.}
  \item Extensive experimental evaluation, comparing our proposed approach to prior work across various metrics and demonstrating its effectiveness in addressing object expansion. Compared to a state-of-the-art baseline, our proposed approach reduces the object expansion by 3.6$\times$ on average with no degradation in standard visual metrics across multiple datasets.
\end{itemize}

\section{Related Work}
\subsection{Diffusion Models}
Diffusion models \cite{NEURIPS2020_4c5bcfec, pmlr-v37-sohl-dickstein15} are a class of generative models that learn the data distribution by learning to invert a Markov noising process. They have gained widespread attention recently due to their training stability and superior performance in image synthesis compared to prior approaches such as generative adversarial networks (GANs). Given a clean image ${x_0}$, the diffusion process adds noise to the image at each step $t$, obtaining a set of noisy images ${x_t}$. Then, a model is trained to recover the clean image ${x_0}$ from ${x_t}$ in the backward process. Diffusion models have produced appealing results on different tasks, e.g., unconditional image generation~\cite{NEURIPS2020_4c5bcfec, pmlr-v37-sohl-dickstein15, JMLR:v23:21-0635, suvorov2021resolution}, text-to-image generation~\cite{Ramesh2022HierarchicalTI, pmlr-v139-ramesh21a, Rombach2021HighResolutionIS, Saharia2022PhotorealisticTD}, video generation~\cite{Ho2022ImagenVH}, image inpainting~\cite{Avrahami2022BlendedLD, Avrahami2021BlendedDF, Lugmayr2022RePaintIU, Nichol2021GLIDETP}, image translation~\cite{Meng2021SDEditGI, Wang2022PretrainingIA, Zhao2022EGSDEUI}, and image editing~\cite{Couairon2022DiffEditDS, Hertz2022PrompttoPromptIE, Kawar2022ImagicTR}.

\begin{figure*}
\begin{center}
\includegraphics[width=0.9\linewidth]{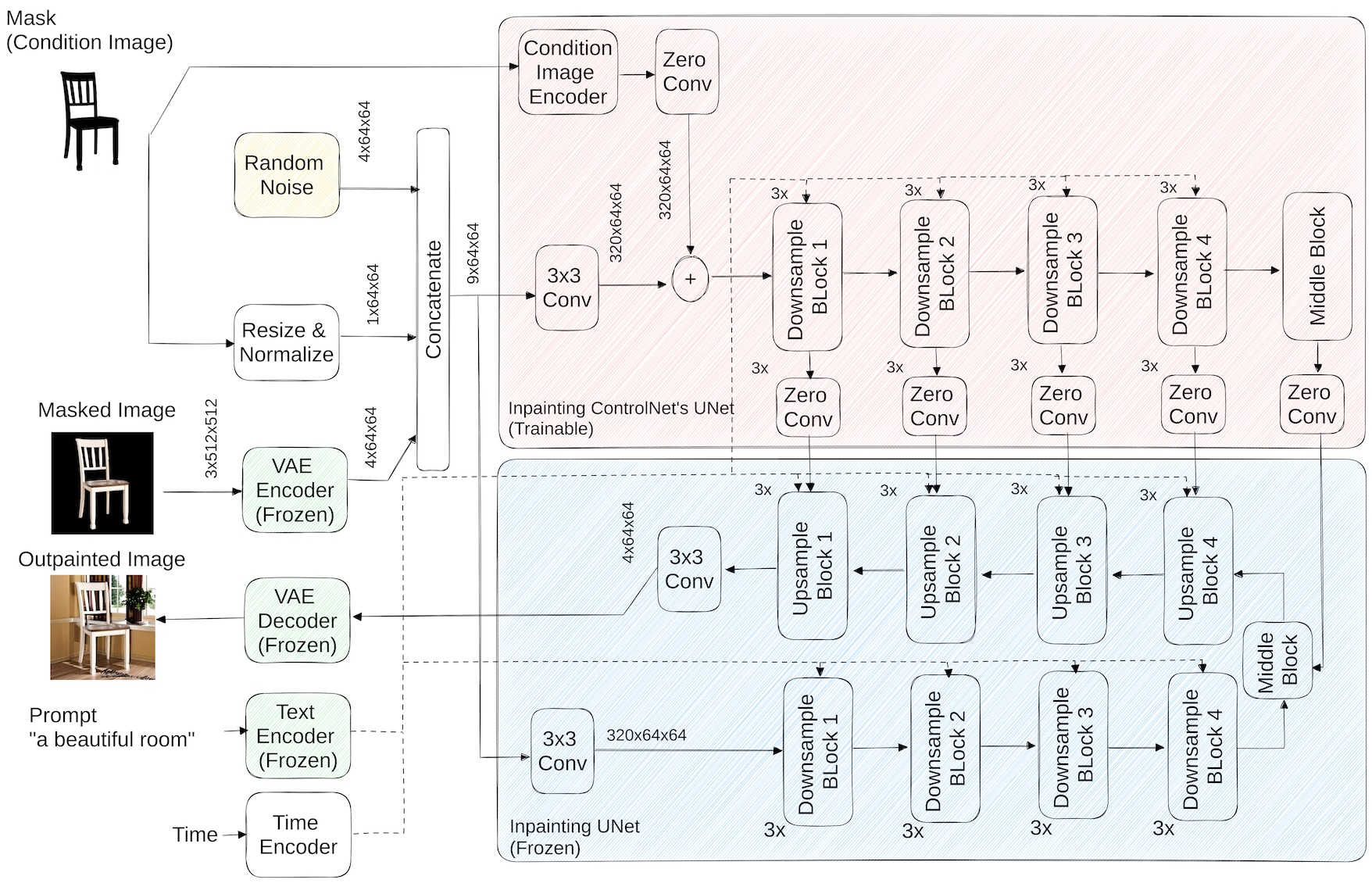}
\end{center}
   \caption{The proposed architecture for salient object outpainting. The original ControlNet architecture only works with text-to-image Stable Diffusion. To make it compatible with text-to-image Stable Inpainting we modified the ControlNet's U-Net architecture to take two extra inputs: 1) mask and 2) masked image. The blue region denotes the frozen Stable Inpainting's U-Net model. The red region includes replicas of the encoder layers of the blue region. The zero convolution outputs from the red region modulate the outputs of decoder layers in the blue region. Initially, during training, the modulation has no effect on the output as the weights of the convolution layer are initialized to zero. Gradually, during training, the nuances of the task of background generation for salient objects will be encoded in modulated values. }
\label{fig:controlnet-inpainting}
\end{figure*}

\subsection{Text-guided Image Inpainting}
Leveraging the recent triumph of diffusion-based text-to-image models, a natural transition from text-to-image creation to text-guided inpainting involves running diffusion with a standard synthesis model, but at each step replacing the portion of the image being generated that is outside the mask with a noised version of the input image. In practice, this approach does not properly condition on the input image, leading to incongruent generations.
GLIDE~\cite{Nichol2021GLIDETP} effectively addresses this issue by using the masked image and mask as direct conditioning inputs to the diffusion model.
Blended Diffusion~\cite{Avrahami2021BlendedDF} promotes the alignment of the final output with the text prompt through the use of a CLIP-based score \cite{Radford2021LearningTV}. The Repaint method \cite{Lugmayr2022RePaintIU} resamples during each retrograde step, yet lacks support for text input. PaintByWord \cite{Bau2021PaintBW} creates an alliance between a large-scale generative adversarial network (GAN) and a complete-text image recovery network, facilitating multi-modal image editing; however, the GAN structure restricts specific modifications to regions indicated by the mask. TDANet \cite{Zhang2020TextGuidedNI} introduces a dual attention mechanism that uses text features related to the masked area by contrasting the text with the original and noised image. SmartBrush \cite{Xie2022SmartBrushTA} proposes a diffusion-based model for completing a missing region with an object using text and shape guidance. None of the prior arts study the task of background generation for salient objects using diffusion models. 

\section{Salient Object Outpainting}
Here, we introduce our proposed model architecture for salient object outpainting. We use Stable Inpainting 2.0 (SI2) as a base model and add the ControlNet model on top to adapt it to the salient object outpainting task. We explain each component of our model in the following subsections.

\subsection{Stable Inpainting}

The training of Stable Diffusion (SD)~\cite{Rombach2021HighResolutionIS}, a text-to-image diffusion model, involved billions of images. The main component of the model is a denoising U-Net, which itself consists of an encoder, a middle block, and a decoder, with skip connections between the encoder and decoder blocks. The U-Net is composed of 25 blocks, divided into 12 symmetric blocks for each encoder and decoder, plus a middle block. There are 25 blocks in total, with 8 being down-sampling or up-sampling convolution layers, and the remaining 17 being main blocks containing two transformer layers and four residual layers each. Each transformer layer contains several cross-attention and/or self-attention mechanisms. CLIP is the source of text embeddings while sinusoidal positional encoding is used for diffusion time steps.

SD has been shown to be a competent and versatile text-to-image generative model. Stable Diffusion \texttt{v2-base}~\cite{stablediffusiontwobase}, referred to as SD2, was initially trained for 550k steps at 256$\times$256 pixel resolution on a subset of LAION-5B \cite{schuhmann2022laion5b} with aesthetic score of 4.5 or higher. SD2 differs from previous versions due to its subsequent training on a dataset with at least 512 x 512 pixel resolution, resulting in more detailed and visually appealing images.
Stable Diffusion \texttt{v2-inpainting}~\cite{stableinpaintingtwo}, referred to here as SI2, is built on top of SD2 and trained for an additional 200k steps.
The training process incorporates the mask-generation approach introduced in LaMa~\cite{suvorov2021resolution}, while adding the latent VAE representations of the masked image as conditioning inputs. We select SI2 as the base model in this work because SI2 already has outpainting capabilities and provides a better initialization compared to SD2. 

\subsection{ControlNet for Stable Inpainting}
ControlNet \cite{zhang2023adding} is a neural network architecture to control the output of existing text-to-image diffusion models by enabling them to support additional input conditions. We adapt ControlNet's architecture to text-to-image \textit{inpainting} diffusion models as they provide a good initialization point for outpainting tasks. This adaptation requires adding extra inputs to ControlNet: (i) a masked image, which contains the pixel values of the salient object; (ii) a binary mask in which \textbf{1}s are the pixels to fill in and \textbf{0}s are the pixels to keep from the salient object in the masked image. Figure~\ref{fig:controlnet-inpainting} shows the architecture of the proposed salient object outpainting. We employ the ControlNet architecture on top of the Stable Inpainting 2.0 (SI2) model.

To enable computationally efficient training, SD applies a pre-processing technique akin to VQ-GAN \cite{esser2020taming} where the entire collection of $512 \cross 512$ images is transformed into smaller $(64 \cross 64 \cross 4)$ \textit{latent images}. To match the convolution size, it is necessary to convert image-based conditions to a $64 \cross 64 \cross 4$ feature space in the ControlNet architecture. The image-space condition is encoded into feature maps with a tiny neural network comprising of four convolution layers. The network uses $4\times4$ kernels and $2\times2$ strides, ReLU activations, and channel dimensions of 16, 32, 64, and 128 (respectively for each of the four convolution layers) and is initialized with Gaussian weights. This network is trained jointly with the ControlNet model and later passed to the U-Net model. Training the ControlNet is computationally efficient as the original weights of the UNet are locked, and only the gradients from the UNet's decoder are required so we need not compute gradients for the original UNet's encoder. 
As only the encoder layers are copied to ControlNet, the cost of running the decoder layers is avoided. Specifically, in the forward pass, we must operate a lightweight mask encoder, two U-Net encoders, one U-Net decoder, and a few zero convolution layers. 
We observe a 33\% increase in runtime and a 25\% increase in GPU memory consumption on V100 GPUs when fine-tuning the whole SI2's U-Net, relative to training the ControlNet.

\begin{figure*}[t]
\begin{center}
\includegraphics[width=0.9\linewidth]{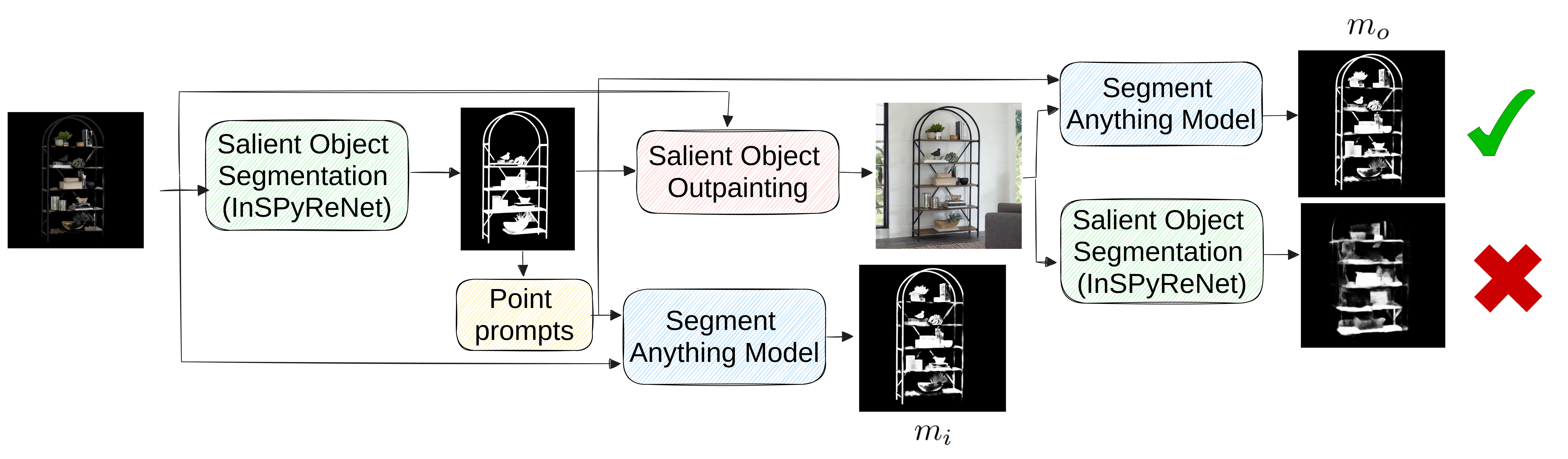}
\end{center}
   \caption{Pipeline for computing salient object masks of the original image ($m_i$) and the outpainted image ($m_o$) to measure object expansion. We found that existing salient object segmentation (SOS) models underperform on synthetic images, but the Segment Anything Model (SAM) works robustly. Therefore, we (i) obtain the salient object mask of the original image using the SOS model, (ii) sample random points from the mask, and (iii) pass sampled point coordinates as the input point prompt to SAM to obtain the salient object mask $m_o$. We obtain a new salient mask from SAM for the original image ($m_i$) as well for an apples-to-apples comparison with $m_o$. }
\label{fig:sam-segment-salient}
\end{figure*}

In the ControlNet's U-Net, the downsampling blocks (i.e., encoder) and middle block are copied from SI2 and initialized using the same weights. The U-Net in the SI2 model takes two additional inputs besides the random noise $(4 \cross 64 \cross 64)$ by concatenation across the channel dimension: (i) binarized mask $(1 \cross 64 \cross 64)$, and (ii) encoded masked image $(4 \cross 64 \cross 64)$. The resulting latent input $(9 \cross 64 \cross 64)$ is then passed to a $3 \times 3$ convolution layer which outputs a tensor of size $320 \cross 64 \cross 64$. The input to the ControlNet's U-Net is the encoded condition image (i.e., salient mask) which is encoded using a convolutional encoder followed by a \emph{zero convolution} layer.

The ControlNet uses several zero convolution layers to modify the U-Net decoder outputs gradually. Mathematically, we are given the feature map $x \in R^{h \cross w \cross c}$ with $\{h, w, c\}$ being height, width, and channel numbers, a U-Net encoder block $E(.;\Theta_e)$ with a set of parameters $\Theta_e$, and a U-Net decoder block $D(.;\Theta_d)$ with a set of parameters $\Theta_d$. We denote the zero convolution operation as $\mathcal{Z}(.;\Theta_z)$. The structure of the ControlNet we are using is defined by:
\begin{equation} \label{controlnet_formula}
\begin{split}
    y = D(d; \Theta_d) + \mathcal{Z}\left(E(e; \Theta_e), \Theta_z \right)
\end{split}
\end{equation}
\noindent where $y$ becomes the output of a decoder layer modulated by the ControlNet structure. As the parameters of a zero convolution layer are initialized as zeros, in the first gradient descent step, we have $\mathcal{Z}(x;\Theta_z) = \mathbf{0}$, which means the original output of the decoder layer does not change. As a result, all the inputs and outputs of both trainable and frozen copies of the U-Net model are not changed, as if the ControlNet did not exist. When the ControlNet structure is applied to some layers before any gradient descent step, it will not influence the intermediate features.

\begin{table*}[!htbp]
\centering

\begin{tabular}{|l|l|ccccc|}
\hline
Dataset &
Model &
FID $\downarrow$ &
LPIPS $\downarrow$ &
CLIP Score $\uparrow$ &
Obj. Similarity $\uparrow$ & 
Obj. Expansion $\downarrow$ \\
\hline

\multirow{5}{*}{ImageNet-1k} &
Blended Diffusion &
31.63 &
0.41 &
24.95 &
0.49 &
0.21 \\
 &
GLIDE &
26.35 &
\textbf{0.28} &
24.82 &
0.62 &
0.18 \\
 &
Stable Diffusion 2.0 &
16.90 &
0.38 &
\textbf{27.46} &
0.56 &
0.15 \\
 &
Stable Inpainting 2.0 &
10.56 &
0.34 &
27.21 &
0.63 &
0.12 \\
\cline{2-7}
&
SI2 + ControlNet (ours) &
\textbf{8.56} &
0.32 &
26.34 &
\textbf{0.69} &
\textbf{0.04} \\
\hline
\hline

\multirow{5}{*}{ABO} &
Blended Diffusion &
30.13 &
0.36 &
25.70 &
0.75 &
0.25 \\
 &
GLIDE &
25.67 &
\textbf{0.19} &
26.17 &
0.80 &
0.26 \\
 &
Stable Diffusion 2.0 & 
9.58 &
0.31 &
\textbf{28.45} &
0.72 &
0.18 \\
 &
Stable Inpainting 2.0 & 
9.31 &
0.28 &
28.10 &
0.80 &
0.10 \\
\cline{2-7}
&
SI2 + ControlNet (ours)&
\textbf{5.93} &
0.27 &
27.74 &
\textbf{0.83} &
\textbf{0.04}  \\
\hline
\hline

\multirow{5}{*}{COCO} &
Blended Diffusion &
30.88 &
0.43 &
23.82 &
0.40 &
0.21 \\
 &
GLIDE &
25.96 &
0.37 &
24.40 &
0.48 &
0.13 \\
 &
Stable Diffusion 2.0 & 
18.89 &
0.42 &
\textbf{27.51} &
0.47 &
0.17 \\
 &
Stable Inpainting 2.0 & 
11.35 &
0.38 &
27.25 &
0.52 &
0.12 \\
\cline{2-7}
&
SI2 + ControlNet (ours) &
\textbf{9.38} &
\textbf{0.36} &
26.37 &
\textbf{0.57} &
\textbf{0.04} \\
\hline
\hline

\multirow{5}{*}{DAVIS} &
Blended Diffusion &
29.98 &
0.48 &
22.14 &
0.52 &
0.12 \\
 &
GLIDE &
24.78 &
0.40 &
24.02 &
0.54 &
0.07 \\
 &
Stable Diffusion 2.0 & 
20.69 &
0.44 &
\textbf{28.14} &
0.56 &
0.16 \\
 &
Stable Inpainting 2.0 & 
11.77 &
0.39 &
28.10 &
0.64 &
0.06 \\
\cline{2-7}
&
SI2 + ControlNet (ours) &
\textbf{8.70} &
\textbf{0.37} &
27.62 &
\textbf{0.69} &
\textbf{0.01} \\
\hline
\hline

\multirow{5}{*}{Pascal} &
Blended Diffusion &
30.10 &
0.45 &
24.33 &
0.48 &
0.12 \\
 &
GLIDE &
26.95 &
\textbf{0.30} &
24.58 &
0.55 &
0.14 \\
 &
Stable Diffusion 2.0 & 
18.83 &
0.40 &
\textbf{27.41} &
0.50 &
0.14 \\
 &
Stable Inpainting 2.0 & 
11.26 &
0.36 &
27.30 &
0.56 &
0.10 \\
\cline{2-7}
&
SI2 + ControlNet (ours) &
\textbf{8.28} &
0.34 &
26.39 &
\textbf{0.59} &
\textbf{0.03} \\
\hline
\end{tabular}
\caption{Evaluation results of text-guided salient object outpainting. Our proposed approach (SI2 + ControlNet) reduces object expansion relative to SI2 by 3.6$\times$ on average, while also surpassing SI2 on the visual metrics (FID, LPIPS). }
\label{results_table}
\end{table*}

\noindent\textbf{Training.}
Image diffusion models learn to progressively denoise images to generate samples. The denoising process can happen in pixel space or a \textit{latent} space encoded from training data. SD uses latent images as the training domain. Given an image (or latent image) $x_0$, diffusion algorithms progressively add noise to the image and produce a noisy image $x_t$, with $1 \leq t \leq T$ being the number of timesteps for which the noise is added. When $t$ is large enough, the image approximates pure noise. Given a set of conditions including timestep $t$, text prompts $c_t$, as well as a task-specific condition (i.e., the salient mask) $c_f$, image diffusion algorithms learn a network $\epsilon_\theta$ to predict the noise, $\epsilon_t$, added to the noisy image $x_t$ with 
\begin{equation} \label{controlnet_loss}
\begin{split}
    \mathcal{L} = \mathbb{E}_{t \sim [1, T], x_0, c_t, c_f, \epsilon_t} \left\Vert \epsilon_t - \epsilon_\theta(x_t, c_t, c_f, t) \right\Vert_2^2
\end{split}
\end{equation}
where $\mathcal{L}$ is the overall learning objective of the entire diffusion model which can be directly used in fine-tuning as well. As we are applying classifier-free guidance (CFG) \cite{Ho2022ClassifierFreeDG}, we randomly drop 10\% of text guidance during training to not drift away from the learned unconditional image generation. Text dropping also facilitates ControlNet's capability to recognize semantic contents from the salient mask.

\subsection{Measuring Object Expansion}
\label{sec:expansion}
The primary limitation of text-guided diffusion models for salient object outpainting tasks is their inability to preserve object boundaries. To assess a method for handling the issue of object expansion, we need a way to measure the error quantitatively. To avoid the need for expensive human labeling, we initially attempted to use salient object segmentation (SOS) models to generate salient masks for both the input (the salient object on a blank background) and output (outpainted) images. The SOS models we experimented with~\cite{10.1007/978-3-031-26293-7_16,qin2022highly} were observed to have extremely poor performance on outpainted images, which we hypothesize is likely due to a distribution shift. However, we note that the Segment Anything Model (SAM) \cite{Kirillov2023SegmentA} is immune to this issue on outpainted images. While SAM is not an SOS model, it can take a set of positive and negative points as a prompt to segment the objects represented by the positive points, while avoiding those represented by negative points. We randomly pick 10 positive and negative points from the salient mask of the original image obtained using the SOS model InSPyReNet \cite{10.1007/978-3-031-26293-7_16}. The positive (negative) points are inside (outside) the mask. Then, the outpainted image and point prompts are passed to the SAM model to segment the salient object and produce a mask $m_o$. We also obtain a salient mask of the input object-only image $m_i$ using the same process to enable an apples-to-apples comparison between masks. Figure \ref{fig:sam-segment-salient} illustrates the pipeline for obtaining these salient masks. 

Given the masks $m_o$ and $m_i$, a natural measure of object expansion $E$ can be defined as:
\begin{equation} \label{object_expansion_metric_1}
\begin{split}
    E = \textsc{area}(m_{o}) - \textsc{area}(m_{i})
\end{split}
\end{equation}
with \textsc{area} expressed as a percentage of the image.
Because our outpainting models never shrink the salient object, the salient mask area in the outpainted image would ideally always be larger than the salient mask area in the original image, i.e., $\textsc{area}(m_{i}) \leq \textsc{area}(m_{o})$. However, as segmentation models are prone to error,
all pixels in $m_i$ may not be included in $m_o$, leading to an underestimate of the magnitude of the expansion. To account for this, we modify the outpainted mask to include $m_i$ and instead propose the following measure for object expansion:
\begin{equation} \label{object_expansion_metric_2}
\begin{split}
    E = \textsc{area}(m_{o} \cup m_{i}) - \textsc{area}(m_{i})
\end{split}
\end{equation}
With this score, an upper bound exists for the object expansion based on the size of the salient object. The larger the salient object in the original image, the lower the upper bound of expansion. 

\section{Experiments}
We use the following salient object segmentation datasets as training data, with 56k images in total: CSSD \cite{10.1109/CVPR.2013.153}, ECSSD \cite{7182346}, DIS5k \cite{qin2022highly}, DUTS \cite{wang2017}, DUT-OMRON \cite{yang2013saliency}, HRSOD \cite{9008818}, MSRA-10k \cite{ChengPAMI}, MSRA-B~\cite{WangDRFI2017}, and XPIE~\cite{8099951}. As addressed in Section \ref{sec:ablations}, training only on salient object datasets can reduce the diversity of generated backgrounds; for this reason, we also include the training partition of COCO \cite{cocodataset}, which has 118K images. The salient masks for COCO were generated using the state-of-the-art InSPyReNet \cite{10.1007/978-3-031-26293-7_16} salient object segmentation model. To train text-guided diffusion models, we need image captions. We use ground truth captions for COCO and obtain the captions for the salient objects datasets using BLIP-2 \cite{Li2023BLIP2BL}. The proposed architecture is trained on 8 NVIDIA V100 GPUs for 300k iterations with the AdamW \cite{Loshchilov2017DecoupledWD} optimizer using a learning rate of $5e^{-5}$ and batch size of 15 per GPU.

\begin{table}
\centering
\begin{tabular}{|ll|ccc|}
\hline
\multirow{2}{*}{} & \multirow{2}{*}{Model} & \multicolumn{3}{c|}{Prompted Background} \\ 
\cline{3-5}
& & Empty & Likely & Unlikely \\
\hline
\multirow{2}{*}{FID $\downarrow$ }
& SI2 & 12.37 & 11.40 & 18.58 \\
& Ours & \textbf{6.44} & \textbf{7.12} & \textbf{10.50} \\
\hline
\multirow{2}{*}{LPIPS $\downarrow$}
& SI2 & \textbf{0.32} & 0.33 & 0.36 \\
& Ours & \textbf{0.32} & \textbf{0.30} & \textbf{0.33} \\
\hline
\multirow{2}{*}{CLIP Score $\uparrow$}
& SI2 & - & \textbf{25.89}  & \textbf{29.01} \\
& Ours & - & 25.69 & 27.79 \\
\hline
\multirow{2}{*}{Obj. Sim. $\uparrow$ }
& SI2 & 0.59 & 0.70 & 0.59 \\
& Ours & \textbf{0.65} & \textbf{0.72} & \textbf{0.62} \\
\hline
\multirow{2}{*}{Obj. Exp. $\downarrow$}
& SI2 & 0.117 & 0.104 & 0.102 \\
& Ours & \textbf{0.038} & \textbf{0.041} & \textbf{0.044} \\
\hline
\end{tabular}
\caption{Comparison of SI2 and our model given different types of prompts: (i) empty, (ii) a likely setting for the object, and (iii) an unlikely setting. Metrics are averaged over all evaluation datasets. }
\label{table:prompt_types}
\end{table}

\begin{figure}
\includegraphics[width=1.0\columnwidth]{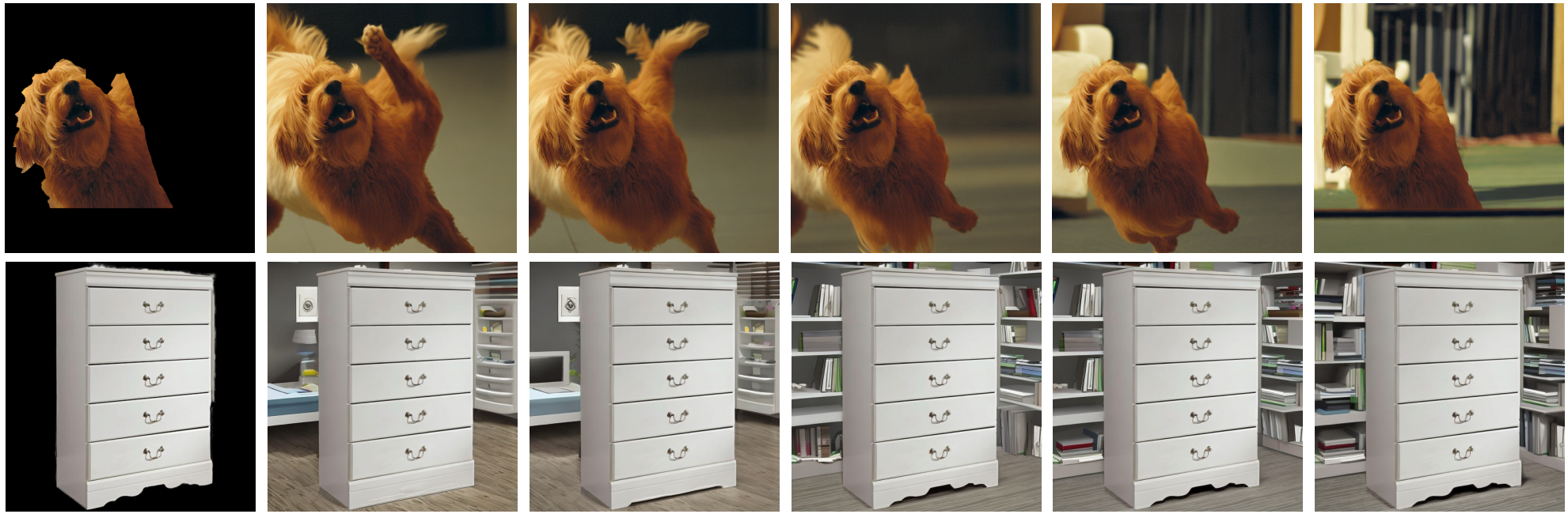}
\begin{tikzpicture}
    \node at (0,0) {\tiny{original image}};
    \node at (1.4,0) {\tiny{$w$\ =\ 0.0}};
    \node at (2.8,0) {\tiny{$w$\ =\ 0.25}};
    \node at (4.2,0) {\tiny{$w$\ =\ 0.5}};
    \node at (5.6,0) {\tiny{$w$\ =\ 0.75}};
    \node at (7.0,0) {\tiny{$w$\ =\ 1.0}};
\end{tikzpicture}
    \vspace{-8pt}
    \caption{Controlling the strength of ControlNet using the adjustable weight $w$ at inference time. With $w=0.0$, objects can expand freely. Setting $w=1.0$ aggressively prevents expansion.}
    \label{fig:weighted_controlnet}
\end{figure}
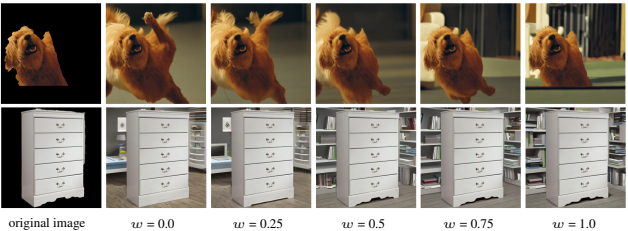

\begin{figure}[!tbp]
\begin{center}
\includegraphics[width=1.0\columnwidth]{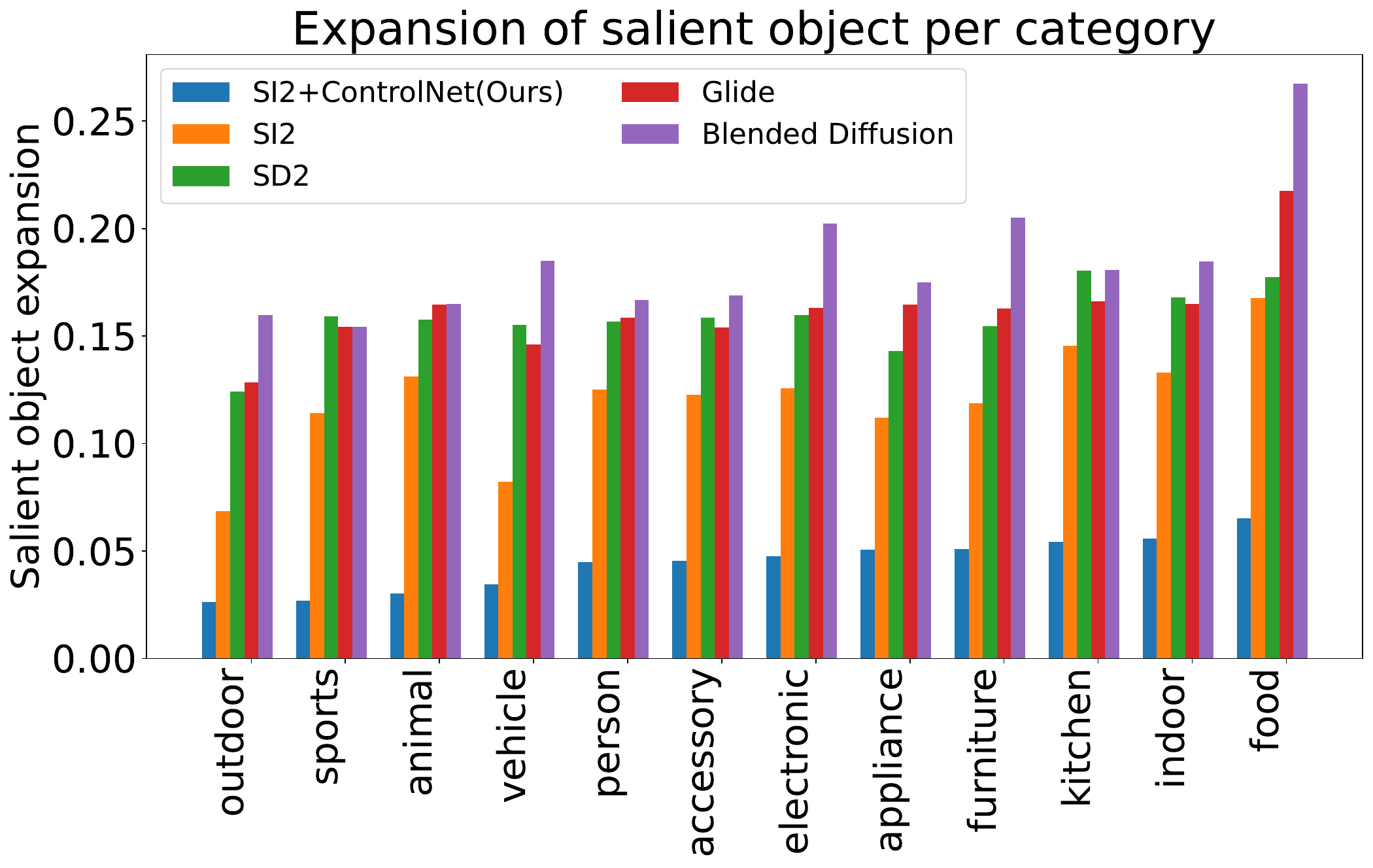}
\end{center}
   \caption{Comparison of salient object expansion across 12 COCO supercategories. The worst expansion scores for each model are observed in indoor settings with fine details. }
\label{fig:expansion-per-category}
\end{figure}

\begin{table*}[!htbp]
\centering

\begin{tabular}{|l|cc|ccccc|}
\hline
Model &
Training Dataset &
Train LPIPS $\downarrow$ &
FID $\downarrow$ &
LPIPS $\downarrow$ &
CLIP Score $\uparrow$ &
Obj. Sim. $\uparrow$ & 
Obj. Exp. $\downarrow$ \\
\hline
\multirow{2}{*}{SD2 + ControlNet} &
SODs &
0.41 &
13.12 &
0.40 &
23.91 &
0.57 &
0.16 \\
&
SODs + COCO &
0.31 &
11.93 &
0.37 &
24.80 &
0.60 &
0.13 \\
\hline
\multirow{2}{*}{SI2 + ControlNet} &
SODs &
0.41 &
9.51 &
0.42 &
25.66 &
0.63 &
0.06 \\
&
SODs + COCO &
\textbf{0.31} &
\textbf{8.17} &
\textbf{0.33} &
\textbf{26.89} &
\textbf{0.68} &
\textbf{0.03} \\
\hline
\end{tabular}
\caption{Comparison of training the ControlNet U-Net initialized with SD2 and SI2 using two training sets: (a) only the salient object datasets (SODs), and (b) SODs plus the COCO training split with segmentation-derived salient object masks. Our results demonstrate that performance measures improve significantly due to (i) initializing with the SI architecture and weights compared to SD, and (ii) adding COCO data to the training set, even without ground truth masks for salient objects. }
\label{sd_si_coco}
\end{table*}

\subsection{Experimental Procedure}
We choose the state-of-the-art image inpainting methods with available code as our baselines: Blended Diffusion~\cite{Avrahami2021BlendedDF}, GLIDE~\cite{Nichol2021GLIDETP}, Stable Diffusion \cite{Rombach2021HighResolutionIS}, and Stable Inpainting~\cite{Rombach2021HighResolutionIS}. Stable Diffusion, Stable Inpainting, and our method support image generation for $512\times512$ sizes, but because Blended Diffusion only supports an image size of $256\times256$, we resize all results to $256\times256$ for fair comparison. We compare these techniques using:
\begin{enumerate}
\item \textbf{Fréchet Inception Distance (FID)} \cite{Heusel2017GANsTB} which evaluates perceptual quality by measuring the distribution distance between the synthesized images and real images. A portion of ImageNet \cite{imagenetdataset} is used as the reference dataset.
\item \textbf{Perceptual Image Patch Similarity (LPIPS)} \cite{8578166} which evaluates the diversity of generated backgrounds by computing the average LPIPS score between pairs of outpainted images for the same salient object image.
\item \textbf{CLIP Score} \cite{hessel-etal-2021-clipscore} which measures the alignment between the text prompt and generated images as the cosine distance between their embeddings using \texttt{CLIP-ViT-L/14}.
\item \textbf{Object Similarity} measures how much the salient object identity is \textit{conceptually} preserved after background generation. This is computed as the cosine distance between the embeddings of the outpainted image and input object-only image using BLIP-2.
\item \textbf{Object Expansion} quantifies the degree of expansion of the salient object in pixel space, as described in Section \ref{sec:expansion}. 
\end{enumerate}
%
We conduct evaluations on five datasets: ImageNet \cite{imagenetdataset}, Amazon Berkeley Objects (ABO) \cite{9879101}, the validation split of COCO \cite{cocodataset}, DAVIS \cite{9008818}, and Pascal \cite{Li_2014_CVPR}. DAVIS and PASCAL already have ground truth salient masks, but we obtain the salient object masks of ImageNet, ABO, and COCO images using InSPyReNet \cite{10.1007/978-3-031-26293-7_16} and discard images in which the salient object occupies less than 5\% of the image area.

\subsection{Results}
The detailed results are presented in Table~\ref{results_table}. Our method reduces object expansion by 3.6$\times$ on average compared to the state-of-the-art SI2. The SI2 model has been trained on the LAION \cite{schuhmann2022laion5b} dataset, which includes billions of web images; however, the web image data may contain non-realistic images such as collages, cartoons, images of text, etc. As we train this model on real image datasets, we obtain improved FID and LPIPS scores across standard datasets, which contain images that also tend to be more realistic. 

After GLIDE, which generates the most diverse backgrounds, our model ranks second in LPIPS by a small margin. However, GLIDE generations perform poorly under FID and CLIP Score and show significant object expansion.
SD2 achieves the highest alignment between text prompt and generation---as measured by CLIP Score---because the generated background is less constrained by the salient object, thus giving the model more freedom to follow the prompt accurately. Our model slightly degrades the CLIP Score of SI2, which may be attributed to the distribution of our training images (67\% COCO) and reliance on BLIP-2 synthetic captions for the salient object datasets in our training corpus. Because these captions can be short and noisy, they may contribute to a decreased adherence to the input text prompt by the trained model. However, our architecture allows controlling the strength of ControlNet at inference time, using an adjustable weight ranging from 0 (no ControlNet) to 1 (full-scale ControlNet). As demonstrated in Figure \ref{fig:weighted_controlnet}, using this feature, one can adjust the amount of control from the ControlNet to different desired levels.

Our approach achieves the highest Object Similarity score, demonstrating that the identity of the salient object is better preserved when expansion is explicitly controlled. A dramatic improvement is seen in the Object Expansion measure from Section \ref{sec:expansion}, with a 3.6$\times$ decrease over SI2, which is ranked second. This improvement can be attributed to both the model architecture and the training data, which effectively address the task of salient object outpainting.  

\subsubsection{Ablation Studies} \label{sec:ablations}

\textbf{Role of text prompts.}
To study the effect of text prompts on the outpainted images, we evaluate our model and SI2 using different types of prompts in Table \ref{table:prompt_types}, including an empty prompt as well as prompts describing likely and unlikely settings for the salient objects. For example, a chair is likely to be found in a room but unlikely to be found in the sky. This is done by using BLIP-2 to caption the salient object image and produce a salient object caption $\sigma$, prompting OpenAI's GPT-4 \cite{openai2023gpt4} with: \textit{``You are a creative and professional photo editor. Question: What is a very/least likely scene for the object described in triple parentheses to be found in? ((($\sigma$))). Answer: The object is very/least likely to be found in''} and then using the API response as the text prompt for outpainting. The results in Table \ref{table:prompt_types} show that FID drops significantly for outpainted images with unlikely backgrounds, while object identity via the Object Similarity score is preserved the most in likely settings. Prompting with implausible backgrounds also leads to a slight decrease in the diversity of the generated backgrounds as shown through LPIPS, but a large increase in prompt alignment via CLIP Score.
We hypothesize that when the object and the prompt are unrelated, the foreground and background become independent during the diffusion process, making them easily distinguishable under these measures. Finally, object expansion does not appear sensitive to the background's naturalness, and our proposed model reduces expansion robustly across different prompt types.

\noindent \textbf{Object expansion across categories.} In Figure \ref{fig:expansion-per-category}, we plot salient object expansion across twelve COCO supercategories. We observe that the ordering of supercategories by the expansion score is fairly similar across the benchmarked models. The highest expansion scores for each model are seen in indoor settings, which tend to contain many fine details and salient objects with less defined dimensions, such as \textsc{food}, \textsc{kitchen}, and \textsc{furniture}. Similarly, the lowest expansion scores occur in outdoor scenes like \textsc{sports} and \textsc{animal} where objects contrast well with the background.

\noindent \textbf{Effectiveness of inpainting models.} 
The original ControlNet architecture was proposed for controlling text-to-image models; however, we have adapted it here to work with text-guided inpainting models. As shown in Table \ref{sd_si_coco}, object expansion with SD2 + ControlNet (a text-to-image model) is higher than that of SI2 + ControlNet (inpainting) because inpainting models already can infill missing image regions, whereas the ControlNet needs to learn this ability from scratch for text-to-image models. SD2 (SI2)'s UNet was used in both the frozen stack and the initialization for the ControlNet encoder stack in the SD2 (SI2) + ControlNet solution.

\noindent \textbf{Effectiveness of expanding the training set.} 
We observed that the background diversity of the salient object datasets is lower than in-the-wild datasets such as COCO. This is also indicated in Table \ref{sd_si_coco} by the LPIPS score for real training images. We added the training split of COCO to our training corpus to improve the diversity of our generated backgrounds. As there is no ground truth segmentation for salient objects in COCO, we generated synthetic salient masks for this data using InSPyReNet \cite{10.1007/978-3-031-26293-7_16}. The results show that including COCO data in training, even with segmentation-derived masks, significantly improves visual and expansion metrics performance.


\section{Conclusions and Future Work}
In this paper, we presented an approach based on diffusion models for generating backgrounds for salient objects without altering their boundaries, as preserving the identity of objects is necessary in applications such as design and e-commerce. We identified the problem of object expansion and provided a measure to capture it. We leave generating backgrounds for non-salient objects as future work because it may require high-quality instance or panoptic segmentation masks. Additionally, future work can explore alternatives to ControlNet such as the T2I-adapter \cite{mou2023t2iadapter}---which modulates the U-Net encoder rather than the decoder---or novel combinations of control architectures for the task of object-aware background generation.


{
    \small
    \bibliographystyle{ieeenat_fullname}
    \bibliography{main}
}


\end{document}